\documentclass[runningheads]{llncs}
\usepackage{graphicx}
\usepackage{algorithm}
\usepackage{algpseudocode}
\usepackage{wrapfig}
\begin{document}
\title{On Reducing Negative Jacobian Determinant of the Deformation Predicted by Deep Registration Networks}
\titlerunning{Reducing folding transformations in Deep Registration Networks}
%
\author{Dongyang Kuang\inst{1}
}
\authorrunning{D. Kuang et al.}
\institute{University of Ottawa, Ottawa, Canada  \\
\email{dykuangii@gmail.com}\\
}
\maketitle              
\begin{abstract}
Image registration is a fundamental step in medical image analysis. Ideally, the transformation that registers one image to another should be a diffeomorphism that is both invertible and smooth. Traditional methods like geodesic shooting approach the problem via differential geometry, with theoretical guarantees that the resulting transformation will be smooth and invertible. Most previous research using unsupervised deep neural networks for registration have used a local smoothness constraint (typically, a spatial variation loss) to address the smoothness issue. These networks usually produce non-invertible transformations with ``folding'' in multiple voxel locations, indicated by a negative determinant of the Jacobian matrix of the transformation. While using a loss function that specifically penalizes the folding is a straightforward solution, this usually requires carefully tuning the regularization strength, especially when there are also other losses. In this paper we address this problem from a different angle, by investigating possible training mechanisms that will help the network avoid negative Jacobians and produce smoother deformations. We contribute two independent ideas in this direction. Both ideas greatly reduce the number of folding locations in the predicted deformation, without making changes to the hyperparameters or the architecture used in the existing baseline registration network. 

\end{abstract}
%
%
%

\section{Introduction}
Image registration is a key element of medical image analysis. 
Most state-of-the-art registration algorithms, such as ANTs \cite{avants2011reproducible}, 
can utilize geometric methods that are guaranteed to produce smooth invertible deformations that are much desired in medical image registration. 
A revolution is taking place in the last couple of years in the application of machine learning methods, especially convolutional neural networks, to this problem. While recent registration networks can make predictions of the nonlinear transformation much faster and obtain registration accuracy comparable to traditional methods, they usually do not have theoretical guarantees on the smoothness and invertibility of their predicted deformations.  

\begin{figure}[tbh]
	\centering
	\includegraphics[width=.75\textwidth]{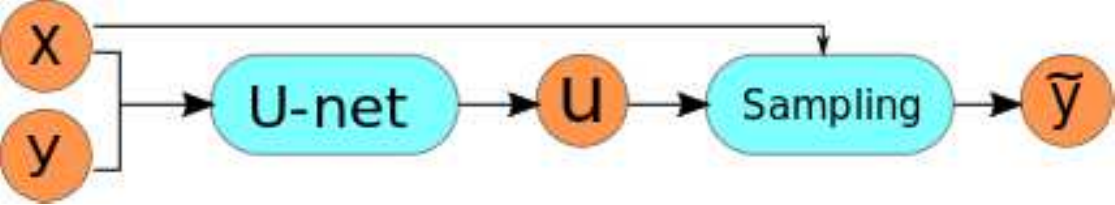}
	\caption{An overview of the registration network usually used for registration. The popular U-net architecture\cite{ronneberger2015u} is used as the deformation unit for generating the displacement field.}\label{fig:baseline}
\end{figure}

Supervised methods, such as in \cite{rohe2017svf,sokooti2017nonrigid,yang2017quicksilver}, learn from known reference deformations for training data -- either actual ``ground truth''  in the case of synthetic image pairs, or deformations computed by other automatic or semiautomatic methods. They usually do not have problems of smoothness, but still relies on other tools such as ANTs running ahead to produce desired transformations. The registration problem is much harder in the setting of unsupervised methods. Most of the unsupervised approaches like \cite{yoo2017ssemnet,li2017non,wang2017scalable,shan2017unsupervised,balakrishnan2018unsupervised} take the idea of spatial transformer (ST) \cite{jaderberg2015spatial}. This spatial transformer used in registration usually consists of two basic function units: a deformation unit and a sampling unit. With input $x$ (source image) and $y$ (target image) stacked as an ordered pair, the deformation unit produces a static displacement field $\bf u : \mathcal{R}^3\to\mathcal{R}^3$. The warped image $\tilde{y}$ is then constructed in the sampling unit by sampling the source image with $\bf u$ via $\tilde{y} = x(Id + \bf u)$, where $Id$ is the identity map. As a summary,  the right action of diffeomorphism $\phi$ on image $x$ is approximated by $\phi \cdot x= x\circ \phi^{-1} \approx x(Id + \bf {u})$. The smoothness constraint on $\bf u$ is usually addressed by regularizing its derivative $D \bf u$. The work \cite{balakrishnan2018unsupervised} is one representative and Figure \ref{fig:baseline} shows the work flow of the idea introduced as above. The whole network is trained so that it minimizes the loss:
$CC(y, \tilde{y}) + \lambda ||D {\bf u}||_{l_2}$,
where $CC$ stands for cross correlation loss and $\lambda$ is a hyperparameter controlling the strength of the regularization.

\begin{figure}[tbh]
	\centering
	\includegraphics[width=.75\textwidth,trim={0cm 0cm 0cm 0cm}, clip]{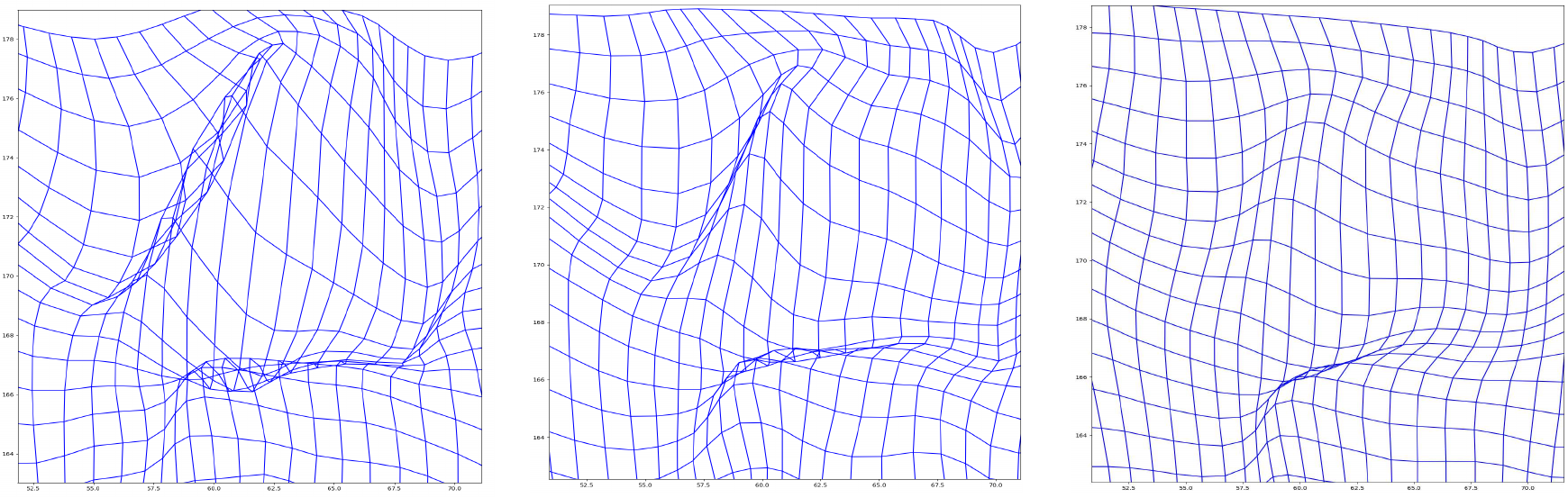}
	\caption{A snapshot of at the same location of the projected warped grid with different regularization strength. From left to right, the network is trained with $\lambda = 1, 2, 4$ separately. }\label{fig:VoxR}
\end{figure}

These work emphasize more on the accuracy and efficiency of registration when compared to classical methods but usually did not put equal emphasis on checking geometric properties such as smoothness, invertibility or orientation preservation for the predicted deformations. Particularly, Jacobian determinant of the predicted transformations i.e. det($D\phi^{-1}) \approx$ det(Id + $D \bf u$) from a neural network can very likely be negative at multiple locations.  This ``folding" issue during prediction may still persist even when one increases regularization strength of $D \bf u$ (see Figure \ref{fig:VoxR}). 
Additionally, this hyper-parameter is usually difficult to set in order to reach a good balance between nice geometrical properties\footnote{In the paper, it will mainly refer to smoothness, invertibility and particularly, transformations has positive Jacobian determinant everywhere.} 
and registration accuracy, since larger $\lambda$ values often cause smaller deformations reducing the accuracy. 
In this paper, we propose two separate ideas: a cycle consistent design and a refinement module focusing specifically on the negative Jacobian determinant issue of unsupervised registration networks. The two ideas inspect the problem from a different angle by altering the training mechanism instead of tuning hyper-parameters or changing the commonly adopted loss functions. 
From our experiment, both methods greatly reduce chances of negative Jacobian determinant in predicted transformations with little sacrifice in registration accuracy.
\section{Proposed methods}
\subsection{Cycle consistent design}
From the mathematical point of view, the transformations used in registration tasks should ideally be diffeomorphisms so that topological properties are not changed during the transformations. In order to approximate the ideal property of invertibility, 
training of the network should also respect this invertibility property.  
In fields such as computer vision, there has already been research such as \cite{zhu2017unpaired} utilizing this idea for better quality control of cross-domain image generation. That work defines two joint cycle consistent loops for better training two separate generative adversarial networks for unpaired image-to-image translation back and forth. We use a related idea in a different setting here for regularizing the predicted static displacement field. This ``cycle consistent loss" idea does not involve new losses but forces the same network to perform a backward prediction trying to recover the input right after it completes the forward prediction.  
\begin{wrapfigure}{R}{5.5cm}
	\centering
	\includegraphics[width=0.4\textwidth, height=80pt]{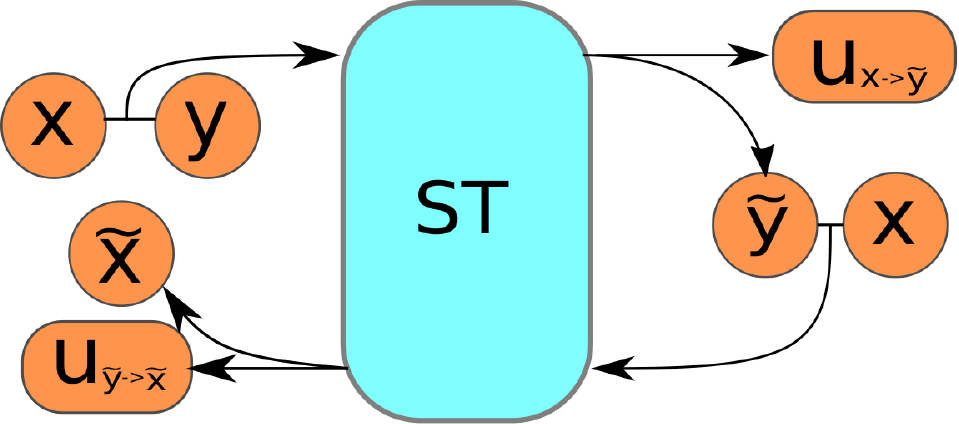}
	\caption{A diagram illustrating the cycle consistent design.}\label{fig:cycle}
\end{wrapfigure}
As seen in Figure \ref{fig:cycle}, the spatial transformer will first predict the warped image $\tilde{y}$ and displacement field ${\bf u}_{x\to\tilde{y}}$ with the stacked source image $x$ and target image $y$.  The predicted target image $\tilde{y}$ (now as source) is then stacked with the original source image $x$ (now as target). They will be fed into the same spatial transformer to produce a reconstruction $\tilde{x}$ for $x$ and corresponding inverted displacement field ${\bf u}_{\tilde{y}\to\tilde{x}}$. The whole network is trained with the cycle consistent loss:
\begin{equation}
CC(y, \tilde{y}) + \lambda ||D {\bf u}_{x\to\tilde{y}}||_{l_2} + CC(x, \tilde{x}) + \lambda ||D {\bf u}_{\tilde{y}\to\tilde{x}}||_{l_2}
\end{equation}

While it is straightforward that this design directly addresses the invertibility of the network, the cycle constraint also contributes to the task of learning a smooth solution in an indirect way: the design regularizes the network by forcing the spatial transformer to learn a solution and its inverse at the same time. This helps the network rule out possible transformations that are not cycle consistent. 
 This design also does not add any additional learnable parameters to the original spatial transformer and can be trained as equally efficient.
 \begin{wrapfigure}{L}{6cm}
 	\centering
 	\includegraphics[width=.5\textwidth]{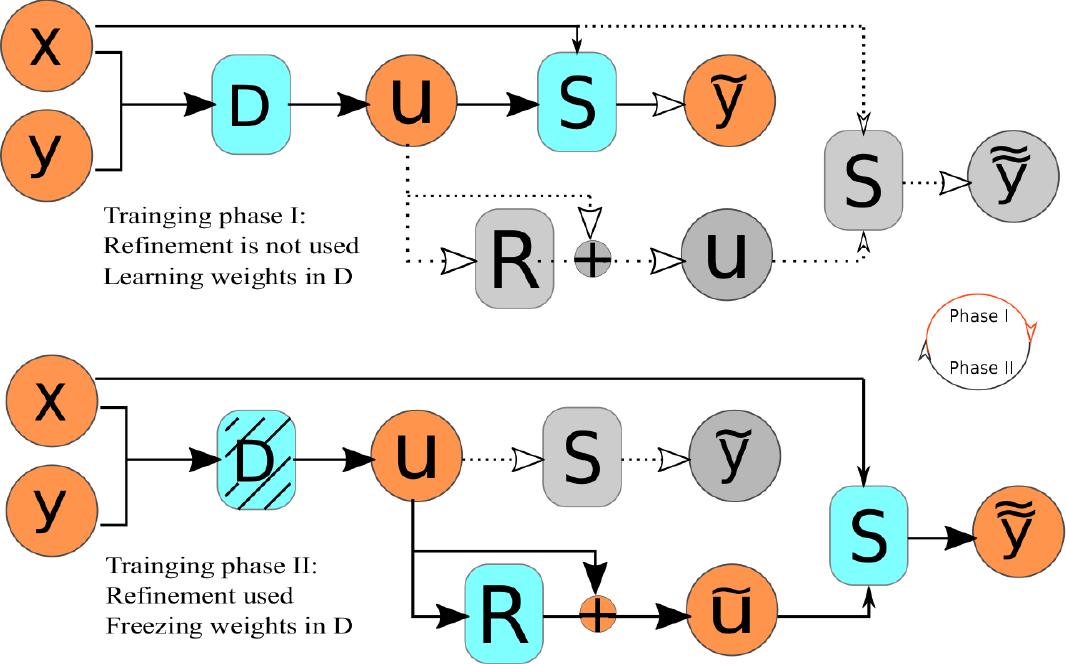}
 	\caption{A diagram showing the activity of the network during training phases. Colored part in the above figure is the baseline network, greyed and dashed part is the attachable refinement network. D: the network producing displacement field. S: the sampling module. R: the network for learning possible corrections needed for smoother field with less ``folding". Training is happening alternatively between these two phases.}\label{fig:refine}
 \end{wrapfigure}
\subsection{Refinement Module}
The other method we propose in this paper is more straightforward by adding a refinement module to the original registration network. This module focuses directly on the task of refining the displacement field. The resulting local smoothness can also contribute to the local invertibility of the transformation through Inverse Function Theorem. Our goal is to separate the task of ``generating a smooth (with less folding) deformation field to warp the source as similar as the target" into two competing subtasks. The generation network (active in phase I) will only focus on improving similarities between warped and target image with mild constraints on the smoothness condition. The following refinement network (active in phase II) with a stronger regularization strength, on the other hand, puts its attention on reducing folding locations by adopting more rigorous smoothness constraints. In order to train both networks together to reach an ideal ``equilibrium" state where the transformation is both accurate and smooth with less foldings, we adopted a alternating training algorithm as seen in Fig \ref{fig:refine}.


\section{Related Work}
To author's best knowledge when completing this paper, \cite{zhang2018inverse} and \cite{goodfellow2014generative} are most relevant research. \cite{zhang2018inverse} designed an inverse consistent network and argued adding an ``anti-folding constraint" to prevent folding in predicted transformation. Different from his work, we did not create new loss in this paper, but focuses on discovering possible training mechanisms that will help regularize the network. The alternating training with refinement model is similar to \cite{goodfellow2014generative}, but our purpose is for regularizing deformation in image transformation instead of image generations. \footnote{The code for the paper is released at https://github.com/dykuang/Medical-image-registration.}
\section{Experiment}
\subsection{Dataset} 
We used MindBoggle101 dataset \cite{klein2012101} for experiments. 
Details of data collection and processing, including atlas creation,
are described in \cite{klein2012101}.
In the present paper, we used brain volumes consisting of the following three named subsets of Mindboggle101:
NKI-RS-22, 
 NKI-TRT-20 and
 OASIS-TRT-20.  
 Each image has a dimension of $182 \times 218 \times 182$, we truncated the marginal reducing the size to $144 \times 180 \times 144$. These images are already linearly aligned to MNI152 space.
We also normalized the intensity of each brain volume to $[0,1]$ by its maximum voxel intensity. 

\begin{figure}[tbh]
	\includegraphics[width=.3\textwidth, height=120pt]{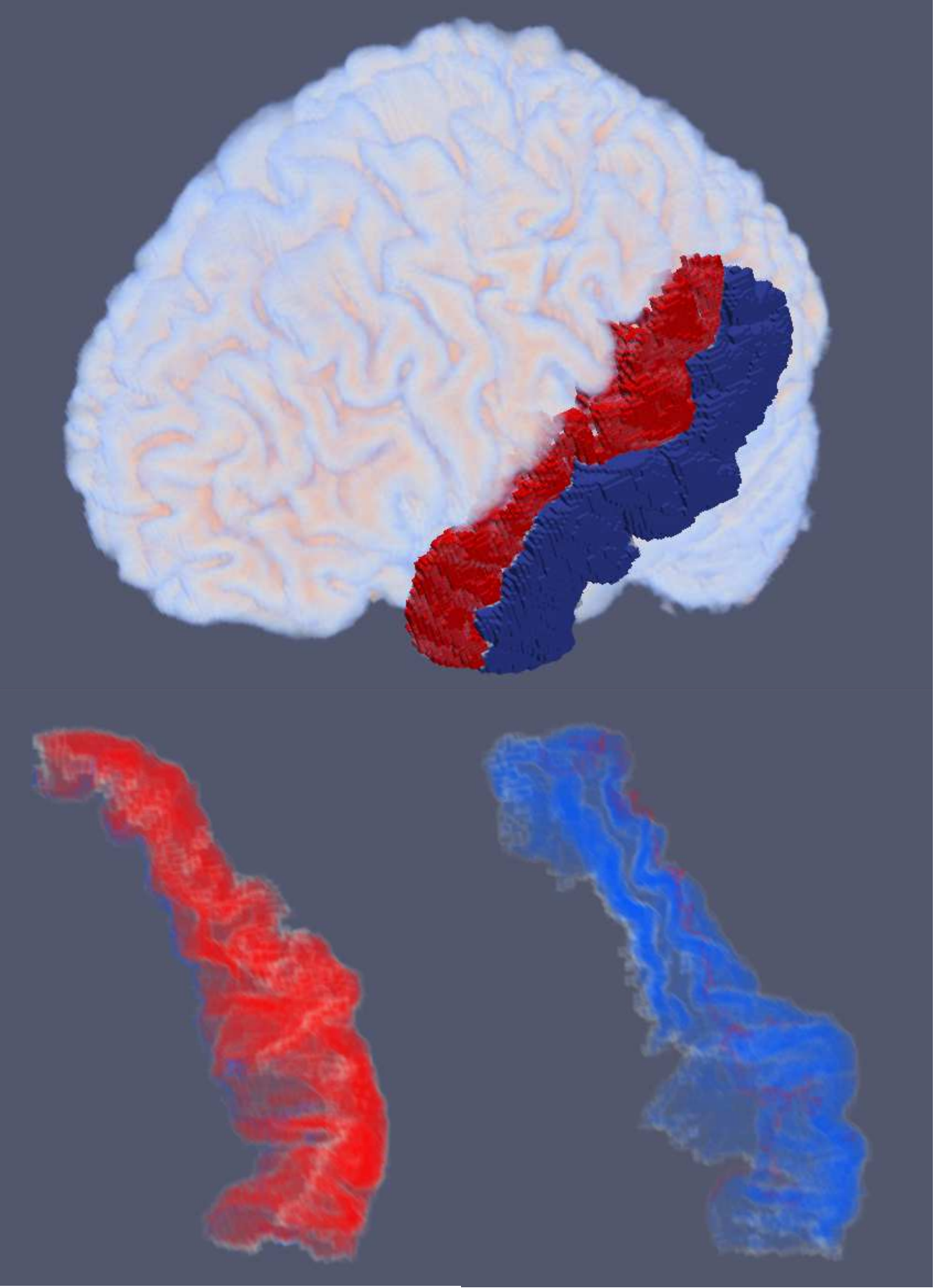}
	\includegraphics[width=.7\textwidth, height=120pt,trim={1cm 1cm 0cm 1cm}, clip]{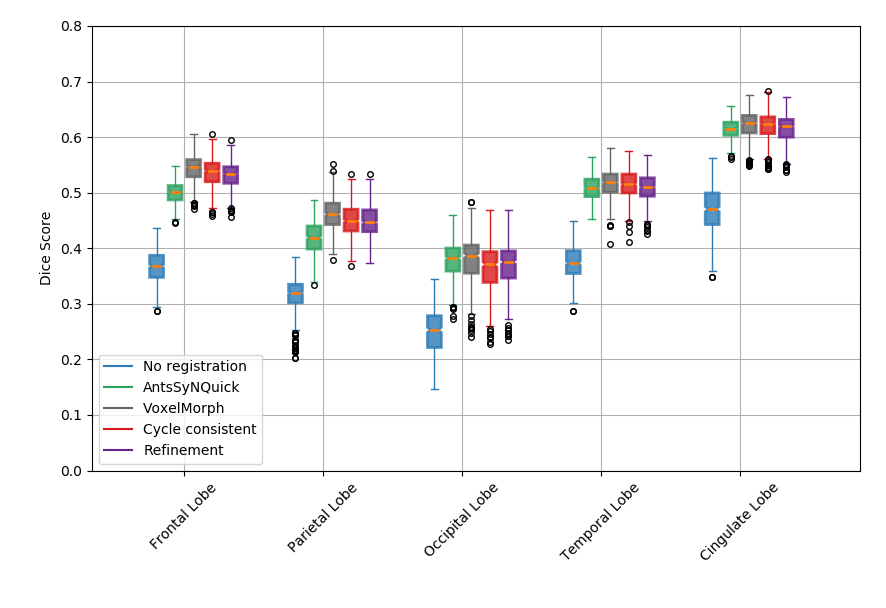}
	\caption{Left: Sample brain volume and 2 labels. Right: Mean dice scores of different methods on selected regions. Each point is the mean dice score averaged over corresponding ROI labels per registration pair instead of over the union of labels in that region. Results from SyNQuick algorithm in the ANTs package are also listed for better interpreting these dice scores, but not for the purpose of comparison. }\label{fig:dice}
\end{figure}

Figure \ref{fig:dice} (left) shows one subject of the dataset with two annotated labels. Labels used in Mindboggle101 data set are cortex surface labels. Their geometrical complexity leads to more challenging registration tasks, especially for neural network approaches. 
In the following experiments, predictions from the original VoxelMorph network  \cite{balakrishnan2018unsupervised} will be used as the baseline network. This baseline network alone, it with cycle consistent design and it with refinement will be compared. The baseline method and the method with cycle consistent design are trained with $\lambda = 1$ and 10 epochs. In the baseline network with refinement, training phase I uses $\lambda=1$ while training phase II takes $\lambda = 4$. Each phase will run 3 epochs continuesly before switching to the other. A total of 4 iterations of this training loop is used. ``Adam" optimizer \cite{kingma2014adam} with learning rate $10^{-4}$ are used for all the three networks. For results showing below, block R in Figure \ref{fig:refine} is simply consisting of two consecutive convolutional layers with Leaky ReLU \cite{xu2015empirical} activation. Better result is possible with a more complex architecture.

We access the accuracy of predicted registration via dice score between ROI labels/masks.  For image pair $(x, y)$, each indexed label $L_x^i $ associated with $x$ will be warped with the predicted deformation $\phi$ from the registration network, dice score is then calculated. 
\begin{equation}
Dice( \,(\phi\cdot L_x^i), L_y^i\, ) = \frac{2|(\phi\cdot L_x^i) \cap L_y^i|}{|\phi\cdot L_x^i | + |L_y^i|}
\end{equation}
This metric on test set (OASIS-TRT-20) is summarized in Fig \ref{fig:dice}.

\begin{figure}[tbh]
	\centering
	\includegraphics[width=\textwidth, height = 90pt]{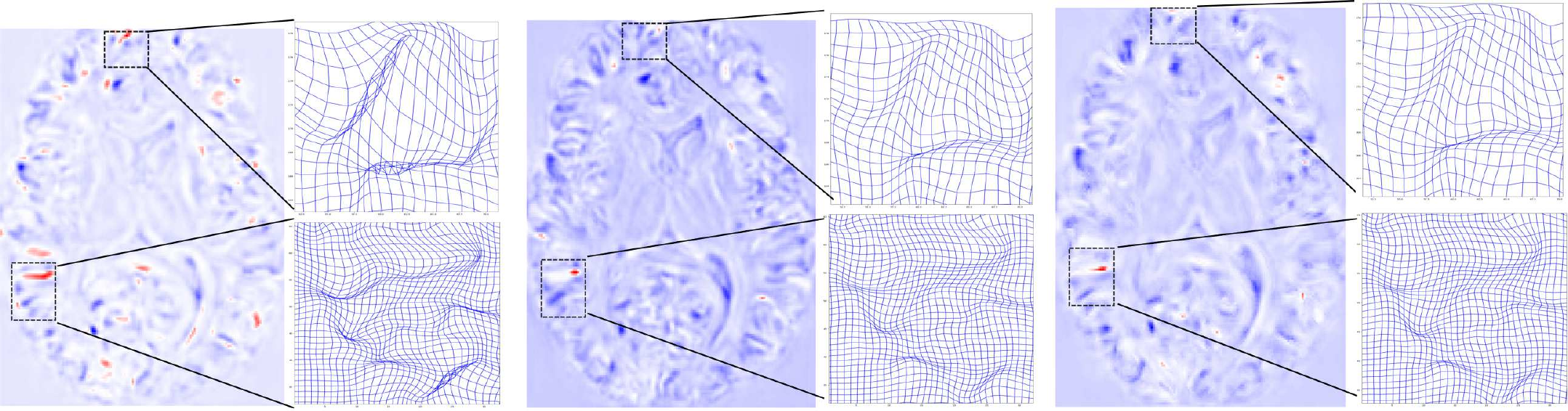}
	\caption{Determinant of Jacobian map and the warped grid projected on the same slice. From left to right:  the baseline VoxelMorph prediction, baseline with cycle consistent design and baseline with refinement module. Locations where the determinants are negative are shown in red.}\label{fig:checkJac}
\end{figure}

\begin{wraptable}{R}{5.5cm}
	\begin{tabular}{l||p{1.5cm}p{1cm}}
		\hline
		Method &   $\mu(\mathcal{P})$ &  $\sigma(\mathcal{P})$ \\
		\hline
		Baseline & 1.97 \% &0.73\%\\
		\hline
		 Cycle Consistent& \textbf{0.13}\%&\textbf{0.05}\%\\
		Refinement & \textbf{0.20}\%&\textbf{0.02}\%\\
		\hline
	\end{tabular}
\caption{Summary of $\mathcal{P}$ with the 3-fold validation.}\label{tab:jac}
\end{wraptable}
The foldings of the deformation is accessed via examining locations where negative Jacobian determinant happen. Let $ \mathcal{P}$ be defined as the percentage of voxel locations where the Jacobian determinant is negative over all voxels $V$, i.e. $\mathcal{P} := \frac{\sum \delta(det(D\phi^{-1}) <0)}{V}.$ The ideal predicted transformation should have this number as small as possible. To better access the general performance of our proposed methods, we perform a 3-fold validation\footnote{Each fold will use 2 of the 3 datasets for forming training set and test on the third. Figure \ref{fig:dice} and figure \ref{fig:checkJac} are from the fold when pairs from OAISIS dataset are used as test. This experiment has 1722 training pairs and 380 test pairs.} with the 3 datasets at hand. We summarize this number from different methods into Table \ref{tab:jac} for comparison. Mean ($\mu (\mathcal{P})$) and standard deviations ($\sigma(\mathcal{P})$) of $\mathcal{P}$ on the test set are recorded. For better visualization, we also put one slice of the Jacobian determinant and the projected warped grid on the same slice in Figure \ref{fig:checkJac}. The transformation for visualization used in the figure is predicted on the pair formed by subject OASIS-TRT-3 (source) and subject OASIS-TRT-8 (target).

Table \ref{tab:jac} suggests there are big differences of the underlying transformation in terms of the measure. From the cross validation results, the baseline method has a mean value of 1.97\% locations where Jacobian determinants are negative. When the cycle consistent design is applied, this value drops to 0.13\%. In other words, more than 90\% of the unsatisfactory locations happening in the baseline prediction are eliminated.  With the refinement module, though not reducing as much as the cycle consistent case, it achieves a mean of 0.20\% with a smaller standard deviation. Surprisingly, Figure \ref{fig:dice} shows that this big improvements in terms of eliminating folding locations did not sacrifice much the registration accuracy measured by dice score.  Figure \ref{fig:checkJac} shows an example of locations of negative Jacobian determinant, this help give an intuitive view of what happened behind the curtain. From the warped grid column, one can clearly see networks with cycle consistent design or refinement module did not change much in locations where the baseline prediction are already smooth but put attentions on foldings and ``unfold" them to produce a much smoother transformation.

%
%

\section{Conclusion}
We contributed two separate ideas for improving the smoothness of deformation when a deep neural network is used for unsupervised registration tasks.  These ideas do not create new kinds of losses but focus on another direction that could bring improvements by adopting different training mechanisms. 
 Both ideas work well in reducing locations that has negative Jacobian determinant when compared to the baseline neural network with little sacrifice of accuracy. They also do not change the baseline registration network and can be used together with other ideas for enhancing smoothness or registration.

%
%
%
{\small
	\bibliographystyle{splncs04}
	\bibliography{tanya}
}
\end{document}